\documentclass[11pt,a4paper]{article}
\usepackage[hyperref]{acl2021}
\usepackage{times}
\usepackage{latexsym}
\usepackage{csquotes}
\usepackage{url}
\usepackage{graphicx}
\usepackage{multicol,array,booktabs}
\usepackage{xcolor}
\usepackage{subfigure}
\usepackage{lipsum}
\usepackage{enumerate}
\aclfinalcopy 

\usepackage{microtype}
\newcommand{\MTL}{{$\langle L_{\textrm{test}}\rangle$}}

\title{Replicating and Extending ``\textit{Because Their Treebanks Leak}'':\\
Graph Isomorphism, Covariants, and Parser Performance}
\author{Mark Anderson \\
 Universidade da Coru\~na, CITIC\\
 Department of CS \& IT \\
  {\tt m.anderson@udc} \\\And
  Anders S{\o}gaard\\
  \textcolor{black}{Dpt. of Computer Science}\\
  \textcolor{black}{University of Copenhagen}\\
  {\tt \textcolor{black}{soegaard@di.ku.dk}} \\\And
  Carlos G{\'o}mez-Rodr{\'\i}guez \\
 Universidade da Coru\~na, CITIC\\
 Department of CS \& IT \\
  {\tt carlos.gomez@udc.es} \\}

\date{}

\begin{document}
\maketitle
\begin{abstract}
\citet{sogaard-2020-languages} obtained results suggesting the fraction of trees occurring in the test data isomorphic to trees in the training set accounts for a non-trivial variation in parser performance. Similar to other statistical analyses in NLP, the results were based on evaluating linear regressions. However, the study had methodological issues and was undertaken using a small sample size leading to unreliable results. We present a replication study in which we also bin sentences by length and find that only a small subset of sentences vary in performance with respect to graph isomorphism. Further, 
the correlation observed between parser performance and graph isomorphism in the wild disappears 
when controlling for covariants. However, in a controlled experiment, where covariants are kept fixed, we do observe a strong correlation. We suggest that conclusions drawn from statistical analyses like this need to be tempered and that controlled experiments can complement them by more readily teasing factors apart.
\end{abstract}

\section{Introduction}
We undertake a replication study of \citet{sogaard-2020-languages} which introduced graph isomorphism (DUG - directed unlabelled graph isomorphism) as a means of explaining differences in parser performance across different treebanks. It measures the ratio of graphs\footnote{Note that in the treebanks used in this paper, namely Universal Dependencies, well-formed trees are enforced.}
in the test set that were also observed in the training data. It is intuitive that this would likely be related to parser performance. 

However, DUG has two important covariants. The size of the training data impacts DUG because the smaller a treebank is, the less likely there will be many crossovers between training and test data. DUG is also tied to the mean sentence length in the test data: smaller sentences are much more likely to have a tree structure already seen in the training, as there are fewer possible trees and the reverse is true for longer sentences, e.g. the number of possible trees for a sentence with 20 tokens is 12,826,228.


\begin{table*}[htbp!]
    \centering
    \small
    \begin{tabular}{l ccc c ccc}
    \toprule
    & \multicolumn{3}{c}{Original} && \multicolumn{3}{c}{10 seeds}\\
    &  CoNLL18 & UDPipe 1.2 & UDPipe 2.0 & &  CoNLL18 & UDPipe 1.2 & UDPipe 2.0 \\
    \midrule
    \textbf{Training size} & 0.014 & 0.100 & 0.060 & &  -0.019 & -0.346 & -0.005 \\
    \textbf{+ DUG} & 0.228 & 0.061 & 0.097 && -0.004 & -0.553 & 0.091 \\
    \textbf{+ \MTL} & 0.195 & 0.169 & 0.146 && -0.007 & -0.370 & 0.140 \\
    \midrule
    \textbf{All}  & -0.078 & 0.157 & 0.086 && -0.413 & -0.138 & 0.106 \\
    \bottomrule
    \end{tabular}
    \caption{Issues with using multivariable linear model and cross-validation (CV) to evaluate explained variance. \textcolor{black}{The first set of columns (Original) uses the exact same settings as the original paper (namely one CV split and the original seed) on the original data (CoNLL18) and the predictions from UDPipe 1.2 and UDPipe 2.0 for the extended data. The DUG explained variance is much smaller for the new data. The second set of columns show the same analysis but averaged over 10 different seeds used for the CV splits. The explained variances are almost all negative, which means the linear fit failed.}}
    \label{tab:replication}
\end{table*}

\section{Related Work}
There is a long history of investigating the causes of variance in parser performance. The effect of training data size on parser performance is well attested \cite{sagae-etal-2008-evaluating,falenska-cetinoglu-2017-lexicalized,strzyz-etal-2019-viable,dehouck-etal-2020-efficient}. Sentence length has also been observed to impact performance \cite{mcdonald-nivre-2011-analyzing}. One likely factor behind this is different sentence lengths having difference dependency distance distributions \cite{ferrer-i-cancho2014the} which in turn affects parsing
as longer dependencies are typically harder to parse \cite{anderson-gomez-rodriguez-2020-inherent,falenska-etal-2020-integrating}. Others have offered explanations based on linguistic characteristics such as morphological complexity \cite{dehouck-denis-2018-framework,coltekin-2020-verification}, part-of-speech bigram perplexity \cite{berdicevskis2018using}, and word order freedom \cite{gulordava-merlo-2016-multi}.

The history of reproduction and replication in NLP is not so well established, with only a few studies in recent years, e.g. on Universal Dependency (UD) parsing \cite{coltekin-2020-verification} and on automatic essay scoring systems \cite{huber-coltekin-2020-reproduction}.

Linear techniques, linear regression models or evaluating correlation coefficients are commonly used for statistical analyses of NLP systems. They have been used to model constituency parser performance \cite{ravi-etal-2008-automatic}, to evaluate what affects annotation agreement \cite{bayerl-paul-2011-determines}, to investigate what impacts statistical MT systems \cite{guzman-vogel-2012-understanding}, what impacts performance on span identifying tasks \cite{papay-etal-2020-dissecting}, and many other examples. Therefore, it is likely that lessons drawn from this replication analysis will be impactful in a broader sense as the conclusions here can be applied in many sub-areas of NLP, namely the sensitive handling of covariants by using partial coefficients, controlled experiments, or signal subtraction; a strong adherence to visualising data; and considering whether the phenomena under consideration are likely to be sensitive to sentence length, as is often the case in NLP, and if so undertaking a sentence-length binning analysis to complement coarser analyses.
\subsection{Original paper}
\textcolor{black}{\citet{sogaard-2020-languages} attempted to explain the difference of parser performance across treebanks by using DUG and also undirected unlabelled graph isomorphism (UUG). Two graphs are isomorphic if there is a renaming of vertices that makes them equal. The first process in calculating DUG (or UUG) is to collect the set of unique graphs that occur in the training data. In the original paper, this set of graphs is referred to as the isomorphisms. Once the training isomorphisms are obtained for a given treebank, the number of graphs in the test data that are members of one of these equivalence classes is counted. The final value is then the proportion of test instances that are isomorphic to the training data. This then gives a value between 0 (all test instances are unique) and 1 (no unique test instances).}  

\textcolor{black}{The analysis was undertaken using a small sample of treebanks that were used at the CoNLL 2018 shared task, using the LAS of the top performing system for each treebank to measure parser performance \cite{zeman-conll18}. The impact DUG (or UUG) has on parsing performance was evaluated by fitting a linear regression to the data with DUG as the control variable. A number of other potential measurements that could explain parser performance were also taken into consideration, but only as alternative explanation and not covariants. The exception to this was using the size of the training data as a covariant. The explained variance and absolute error for each linear regression fit was reported using a three-fold cross-validation. The results suggested that DUG was the most strongly correlated measurement evaluated. We show that this result does not hold up when accounting for covariants, that using cross-validation method with the linear regression is not a robust method for an analysis like this, and that by controlling the main covariants of DUG, we can observe a more trustworthy correlation to parser performance.}
\section{Analysis and results}
We evaluate directed graph isomorphism (DUG) as it was more strongly related to parser performance in the original paper. 

\paragraph{Main covariants} We focus on the two main covariants of DUG: training data size (in sentences) and mean sentence length of the test data, \MTL.
\paragraph{Data and parsers} The data from the original paper consists of 33 UD treebanks, with LAS taken from the respective top performing parser from the CoNLL 2018 shared task \cite{zeman-conll18}. Note that these systems are all variations of the biaffine graph-based parser of \citet{dozat20161}. For replication, we also use a neural transition-based system UDPipe 1.2 \cite{straka2016udpipe}, using UD models 2.4 and UD v2.5 \cite{ud25}, and a neural graph-based system UDPipe 2.0 \cite{straka2018udpipe}, using UD models 2.6 and UD v2.7 \cite{ud27}. This results in 94 treebanks for UDPipe 1.2 and 90 for UDPipe 2.0. The difference is due to issues running the web-based UDPipe 2.0 on larger files.
\subsection{Reproduction and replication}
In the original paper, the analysis focuses on fitting a multi-variable linear regression to the data to control for covariants. However, the models only used training size plus one other variable as features. Further, cross-validation is used so as to avoid over-fitting. While over-fitting isn't directly an issue, the metrics that are typically reported overestimate the variance explained by a linear model, e.g. explained variance, $\eta^2$, or $R^2$ \cite{stats}. Averaging $\eta^2$ over different splits can potentially offset this positive bias but it requires a certain amount of data to be reliable. In Table \ref{tab:replication}, we show the results using the original data from \citet{sogaard-2020-languages}.  The values shown in the left-most column are exact reproductions of the original values. 
Only the value for {\MTL} is different as the original paper appears to have used a normalised value. We also show $\eta^2$ for the linear model using all variables, which is negative, i.e.\ the fit failed.

We next show the results using UDPipe 1.2 and 2.0. While the values for training size on its own and with {\MTL} are similar, the high $\eta^2$ for training size with DUG is no longer observed. This seems to be due to specious results born out of serendipitous splits for the smaller sample from CoNLL 2018. 

We then tested this same procedure using different seeds
to shuffle the cross-validation splits. The results are almost exclusively negative, i.e.\ the linear models failed to fit to the data at all. This further highlights an issue of using this methodology when sample size is small, as the random split can have large impact on the statistical metrics.

\begin{table}[b!]
    \centering
    \small
    \tabcolsep=0.05cm
    \begin{tabular}{lrrrr}
    \toprule
\multicolumn{1}{c}{} & \multicolumn{1}{c}{CoNLL18} & \multicolumn{1}{c}{UDPipe 1.2} & \multicolumn{1}{c}{UDPipe 2.0}  \\
\midrule
\textbf{size} & 0.46 (p$=$0.007) &  0.54 (p$<$0.001) &  0.37 (p$<$0.001)    \\
\textbf{DUG} & -0.13 (p$=$0.458) & -0.13 (p$=$0.213)  &  -0.18 (p$=$0.083)   \\
\textbf{\MTL}& 0.20 (p$=$0.272) & 0.35 (p$=$0.001)  & 0.33 (p$=$0.001)  \\\midrule
\textbf{size} & 0.44 (p$=$0.011) & 0.42 (p$<$0.001) & 0.46 (p$<$0.001)& \\
\textbf{\MTL} & -0.96 (p$<$0.001) & -0.91 (p$<$0.001) & -0.92 (p$<$0.001)  \\
\bottomrule
    \end{tabular}
    \caption{Spearman's $\rho$ for variables with respect to LAS (top) and DUG (bottom).}
    \label{tab:correlations}
\end{table}
\begin{figure*}[htpb!]
    \centering
    \includegraphics[width=0.7\linewidth]{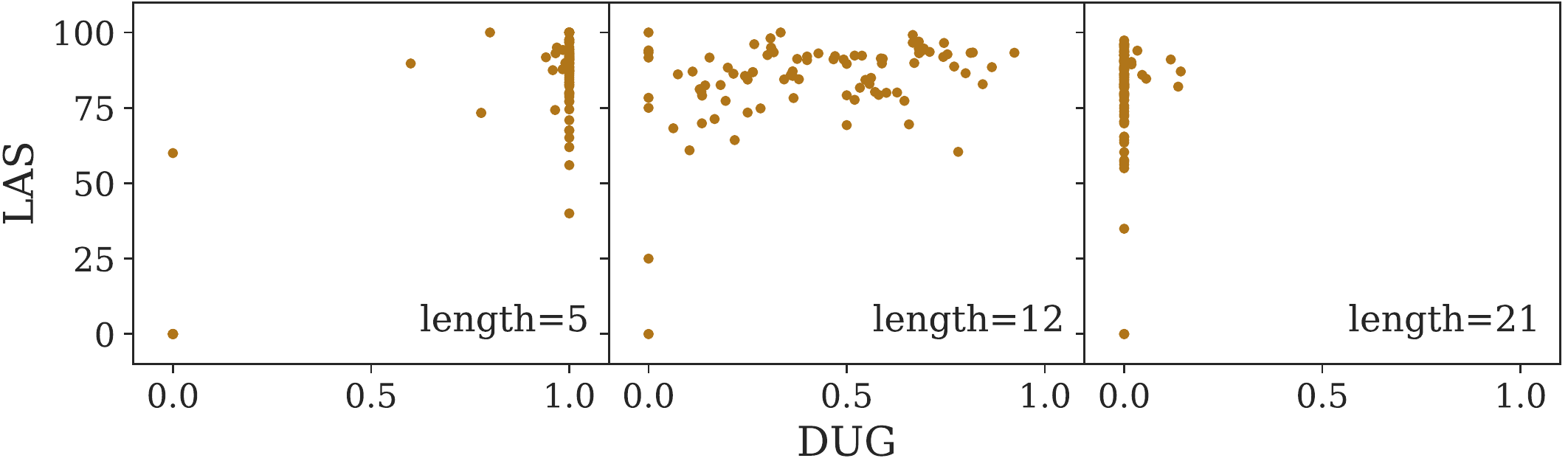}
    \caption{DUG binned wrt sentence length. Values are for UDPipe 2.0 with UD v2.7 for 90 treebanks.}
    \label{fig:example_sentence_length_bins}
\end{figure*}
\subsection{Extending the analysis}
As the linear models performed so poorly, we measured the correlation coefficients (Spearman's $\rho$) for each of the variables with respect to LAS and also the potential covariants with respect to DUG. These are reported in Table \ref{tab:correlations} and we include visualisations of these in Figures \ref{fig:sogaard_scatters} and \ref{fig:udpipe2_scatters} in the Appendix for the CoNLL 2018 data and the UDPipe 2.0 data.
Interestingly, DUG has the highest p-value for all systems, far from statistical significance. However, DUG appears to be strongly correlated to both covariants, especially {\MTL}
with $\rho > 0.9$ and $p < 0.001$ for all datasets and systems. Also of note is that training data size is convincingly correlated to LAS, but based on the linear models it doesn't appear to be predictive of parser performance. Based on this and on the visualisation of the data in Figures \ref{fig:sogaard_scatters} and \ref{fig:udpipe2_scatters} in the Appendix (as well as visualisations of training size vs. LAS in the literature, see \S 2), it seems clear that the relation between these variables is not linear but logarithmic. We show LAS against training data size with a logarithmic scale in Figure \ref{fig:size_logged} in the Appendix.

Table \ref{tab:re_results_logged} shows the results of the limited linear model and cross-validation technique using 10 different seeds as above and using log training size. For these results, the explained variance of the models are all positive and relatively high, that is, the models manage to fit the data unlike in the original setup. This one change offsets the failure of the linear model technique, which is not surprising. However, it seems to suggest that DUG is not a useful feature, as training size with {\MTL} outperforms training size with DUG for all datasets except CoNLL18. And the models which use all features are worse than just using training data size and {\MTL}, with the CoNLL18 model resulting in a negative explained variance, again meaning the fit failed. For CoNLL18, training data size and DUG does outperform the model using \MTL. 

\begin{table}[t!]
    \centering
    \small
    \begin{tabular}{lccc}
    \toprule
    & \multicolumn{1}{c}{CoNLL18} & \multicolumn{1}{c}{UDPipe 1.2} & UDPipe 2.0 \\
    \midrule
        \textbf{log-size}  & 0.055 & 0.319 & 0.126  \\
        \textbf{+DUG} &  0.132 & 0.410 & 0.277 \\ 
        \textbf{+\MTL} &  0.106 & 0.452 & 0.294\\\midrule
        \textbf{All} & -0.184 & 0.412 & 0.229 \\
        \bottomrule
    \end{tabular}
    \caption{Using multivariable linear model and CV to evaluate explained variance with random shuffling (10 splits) and logarithmic transformation of treebank size.}
    \label{tab:re_results_logged}
\end{table}

\subsection{Sentence length binning}
We analyse the relation between test sentence lengths and DUG by binning the data with respect to sentence length. This entails taking each sentence of length $l$ for each treebank, in both the training and test data, and calculating DUG and the corresponding LAS based on these subsets. Figure \ref{fig:example_sentence_length_bins} shows some of these bins (for sentences of length of 5, 12, and 21 tokens) for UDPipe 2.0. A full visualisation of each bin ranging from length 3 tokens to 30 is shown in Figure \ref{fig:dug-binned} in the Appendix. 

DUG is almost exclusively 1.0 for shorter sentences, as can be seen in Figure \ref{fig:example_sentence_length_bins} for sentence length 5. The number of possible directed trees for sentences with less tokens is too small for there not to be crossover: there are only 9 possible unlabelled trees for sentences of length 5 \cite{oeisA000081}. Conversely, for longer sentences, DUG is almost exclusively 0.0 as the number of possible tree structures is considerable (35,221,832 for sentences of length 21).

For a small subset of sentence lengths, ranging from length 9 to 14, there is meaningful spread of values for DUG, with a broadly-speaking linear relation with respect to LAS. 
\begin{table}[b!]
    \centering
    \small
    \begin{tabular}{lcc}
    \toprule
    & UDPipe 1.2 & UDPipe 2.0 \\
    \midrule
    \textbf{LAS} & 0.47 (p$<$0.001) & 0.31 (p$=$0.003)  \\
    \textbf{size} & 0.91 (p$<$0.001)& 0.91 (p$<$0.001) \\
    \textbf{\MTL} & 0.32 (p$=$0.002) & -0.34 (p$=$0.001) \\ \midrule
    \textbf{log-size} &  0.319 & 0.126 \\
    \textbf{+DUG} &  0.331 & 0.147 \\
    \textbf{+\MTL} & 0.452 & 0.294 \\
    \textbf{All} & 0.406 & 0.265 \\ \bottomrule
    \end{tabular}
    \caption{Correlations wrt focused DUG (top) and explained variance (bottom) for focused DUG (sentence lengths 9 to 14) with shuffling for CV (10 seeds).}
    \label{tab:focused_metrics}
\end{table}
Based on this result, i.e.\ that only certain sentence lengths are suitable for using DUG, we considered using a \textit{focused} version of DUG, i.e.\ a variant calculated considering only sentences between length 9 and 14 in the training and test data.
We then analysed how this measurement correlated with parser performance. Table \ref{tab:focused_metrics} shows the correlations for focused DUG with respect to LAS, training size, and {\MTL}. While the correlation between focused DUG and LAS is much higher than for DUG and LAS, this is due to the focused version being much more strongly correlated to training size ($\rho=0.91$ with a p-value less than 0.001 for both datasets) and the correlation with {\MTL} is much diminished. Also, this focused version of DUG improves performance for the linear model when used only with training data size, but {\MTL} improves it much more. Using all 3 is again worse than just using training data size with {\MTL}, however, focused DUG doesn't lower the performance as much as the full variant does.

\subsection{Controlling covariants}
\begin{figure*}[!btp]
    \centering
    \includegraphics[width=0.99\linewidth]{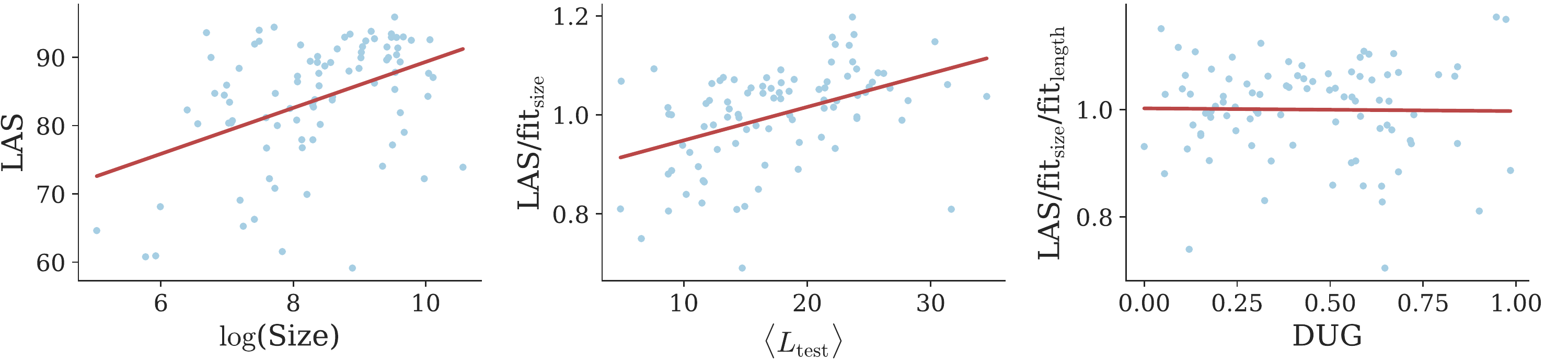}
    \caption{Visualisation of removing background signal associated with covariants of the log of training size (log(Size)) and mean test length \MTL. The spearman's $\rho$ for DUG and LAS is -0.18 (p$=$0.083), for DUG and LAS/bcg$_{\textrm{size}}$ is -0.40 (p$<$0.001) compared to \MTL and LAS/bcg$_{\textrm{size}}$ of 0.465 (p$<$0.001), and finally DUG and LAS/bcg$_{\textrm{size}}$bcg$_{\textrm{Ltest}}$ is 0.01 (p$=$0.926).}
    \label{fig:noise-signal}
\end{figure*}

Having established that DUG does not improve linear models predicting LAS and that DUG is strongly correlated to training treebank size and \MTL, we attempted to find a signal by removing the background signals associated with these variables. We applied a linear fit to the training data size and LAS and then divided the LAS scores by the predicted values of that fit. Then we applied a linear fit to {\MTL} and these \textit{normalised} values and again divided these values out. Finally, we evaluated these \textit{doubly normalised} values against DUG. This process is shown in Figure \ref{fig:noise-signal} for UDPipe 2.0 and the resulting coefficients for UDPipe 1.2 and 2.0 are in Table \ref{tab:noise_correlations} of the Appendix. Removing the signals of the covariants results in a linear fit against DUG with a zero gradient and with a coefficient of 0.01 (p$=$0.926). Removing the variance associated with these covariants effectively removes any signal associated with DUG. 

\begin{table}[b!]
    \centering
    \small
    \tabcolsep=0.05cm
    \begin{tabular}{lrrr}
    \toprule
\multicolumn{1}{c}{} & \multicolumn{1}{c}{CoNLL18} & \multicolumn{1}{c}{UDPipe 1.2} & \multicolumn{1}{c}{UDPipe 2.0}  \\
\midrule
\textbf{DUG} & -0.13 (p$=$0.458) & -0.13 (p$=$0.213)  &  -0.18 (p$=$0.083)   \\\midrule
\textbf{size}& -0.44 (p$=$0.010) & -0.50 (p$<$0.001)  & -0.46 (p$<$0.001)  \\
\textbf{\MTL} & 0.18 (p$=$0.329) & -0.13 (p$=$0.213) & 0.21 (p$=$0.049) \\
\textbf{both} & -0.27 (p$=$0.126) & 0.01 (p$=$0.915) & -0.12 (p$=$0.245)  \\
\bottomrule
    \end{tabular}
    \caption{Partial Spearman's $\rho$ for DUG with covariants.}
    \label{tab:partial_correlations}
\end{table}

To corroborate this background subtraction analysis, we also report the partial coefficients in Table \ref{tab:partial_correlations}. When controlling for both covariants, correlations are small, and p-values very high, for both UDPipe systems. 
CoNLL18 
has a stronger signal, but it is negative (which is the opposite relation one would expect) and has a large p-value.

\subsection{Controlled experiment - fixing covariants}
\begin{figure}[b!]
    \centering
    \includegraphics[width=0.95\linewidth]{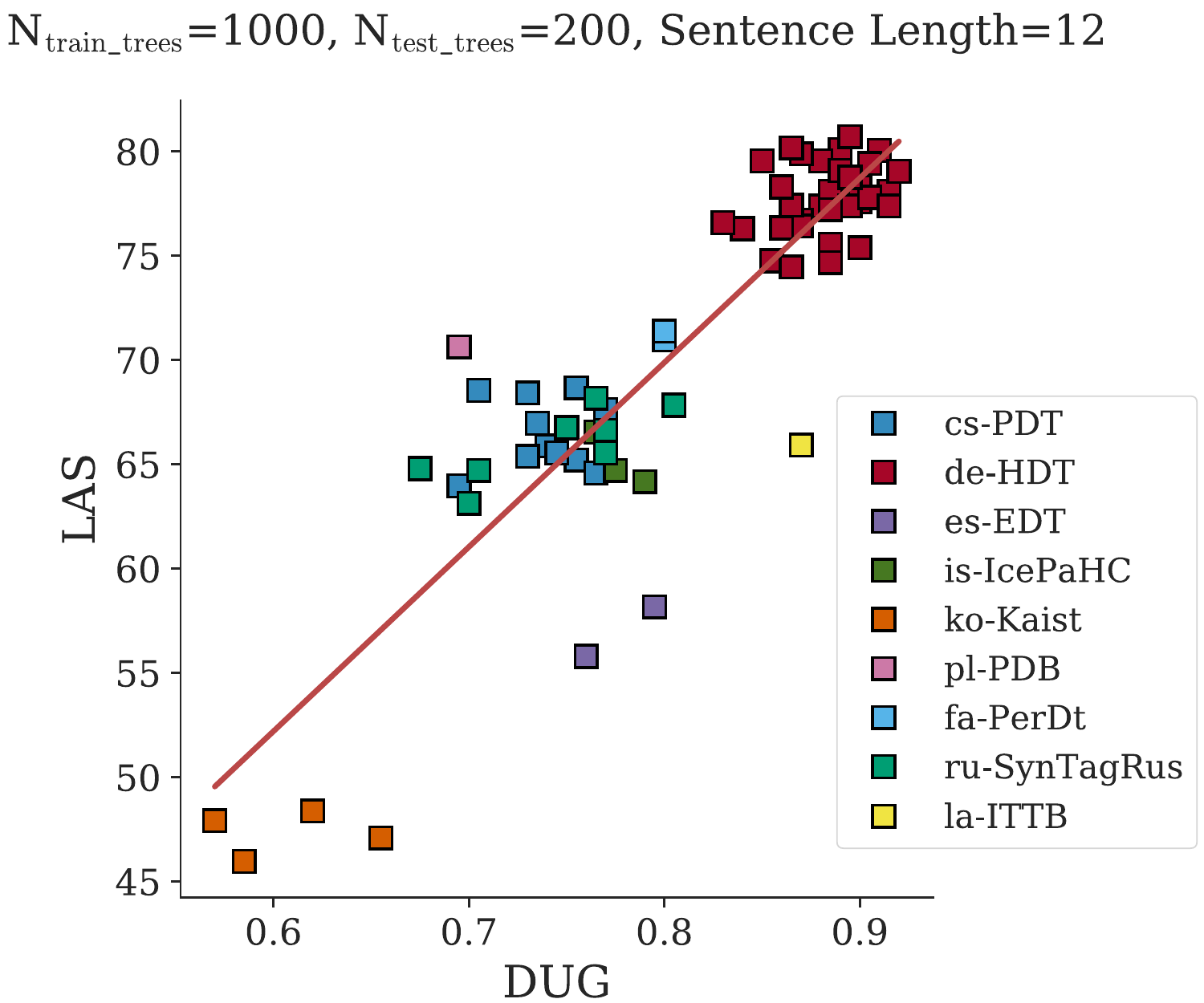}
    \caption{DUG vs LAS for controlled experiment. $\rho=0.82$ (p$<0.001$).}
    \label{fig:controlled}
\end{figure}

We also evaluated DUG's relation to LAS in a controlled experiment where we sampled subsets of treebanks keeping training data size constant and also the sentence length of both training and test data. We trained UDPipe 1.2 models (UDPipe 2.0 is not available beyond using pre-existing models), using standard settings. We were limited to 9 treebanks, 
as we required a reasonable amount of data and using only one sentence length reduces the number of usable treebanks. We combined all of the data for treebanks which had over 1200 sentences of length 12. We then created splits such that a single 1000-sentence training set
was created by randomly sampling sentences. Then a number of 200-sentence test sets were created, generating as many splits as the data allowed for a given treebank. In this way we varied DUG indirectly, but by using different treebanks to sample from we obtained values spanning a reasonable range (0.6 - 0.9). This results in a Spearman's $\rho$ of 0.82 (p$<$0.001) and is visualised in Figure \ref{fig:controlled} in the Appendix. So in this rigid context, we do observe a very strong correlation between DUG and LAS, echoing the analysis from the sentence-length binning procedure.

\section{Conclusion}
With this case study we have shown the value of replicating analyses in NLP. Our analysis has shown that the original results were unreliable and it has highlighted methodological issues the original analysis had. Also, the results regarding the methodology presented here (i.e.\ the need to visualise and evaluate correlations before considering linear regression techniques, the potential sensitivity to sentence length of measurements used in NLP statistical analyses, the need to control for all covariants and evaluate their impact using partial coefficients at the very least, and finally that using controlled experiments can help better evaluate the impact of specific measurements and can complement statistical analyses) will likely be useful for other statistical analyses in different areas of NLP.

\section*{Acknowledgements}
MA and CGR received funding from the European Research Council, under the EU's Horizon 2020 research and innovation programme (FASTPARSE, grant agreement No 714150), from MINECO (ANSWER-ASAP, TIN2017-85160-C2-1-R), from Xunta de Galicia (ED431C 2020/11), and from 
CITIC, funded by Xunta de Galicia and the European Union (ERDF - Galicia 2014-2020 Program), by grant ED431G 2019/01. AS received funding from a Google Focused Research Award.
\bibliography{emnlp2018}
\bibliographystyle{acl_natbib}
\appendix
\section{Appendix}
\label{sec:appendix}
\begin{figure}[bp!]
    \centering
    \includegraphics[width=0.91\linewidth]{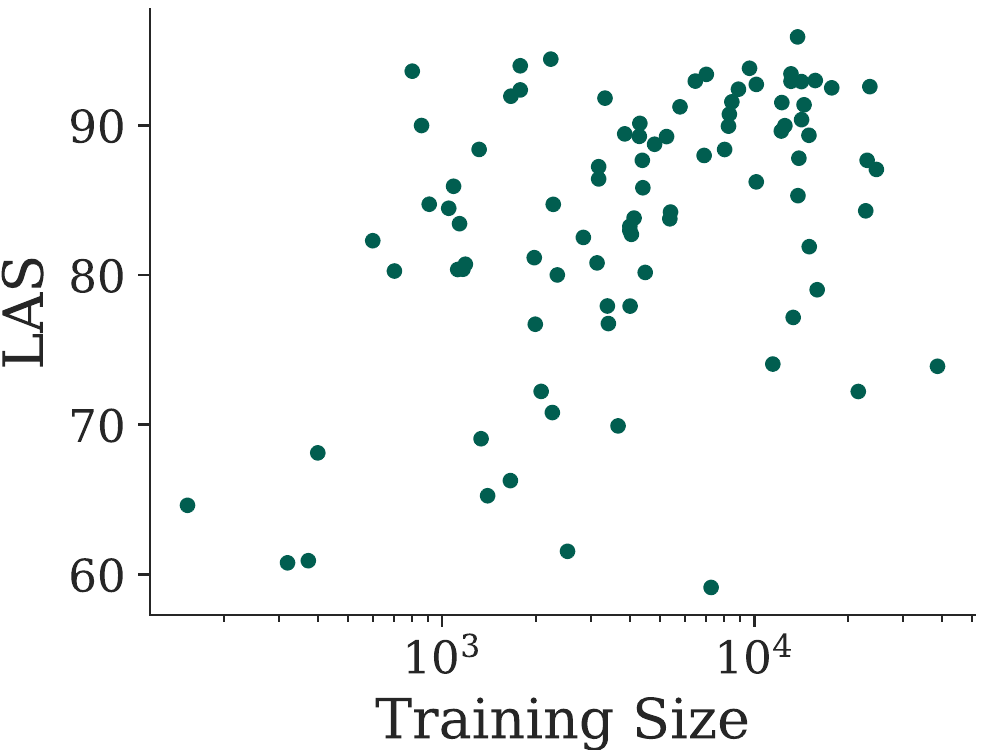}
    \caption{LAS with respect to training set size, in logarithmic scale, for UDPipe 2.0 and UD v2.7.}
    \label{fig:size_logged}
\end{figure}
\textcolor{black}{The appendix mainly consists of visualisations corresponding to the statistical analyses described in the main body. Some additional information is given to supplement the main analyses in Tables \ref{tab:partial_correlations-focused} and \ref{tab:noise_correlations} which give the correlations for the focused DUG analysis and the background removal process, respectively.}

\begin{figure*}[t!]
\centering
\begin{subfigure}
    \centering
    \includegraphics[width=0.95\linewidth]{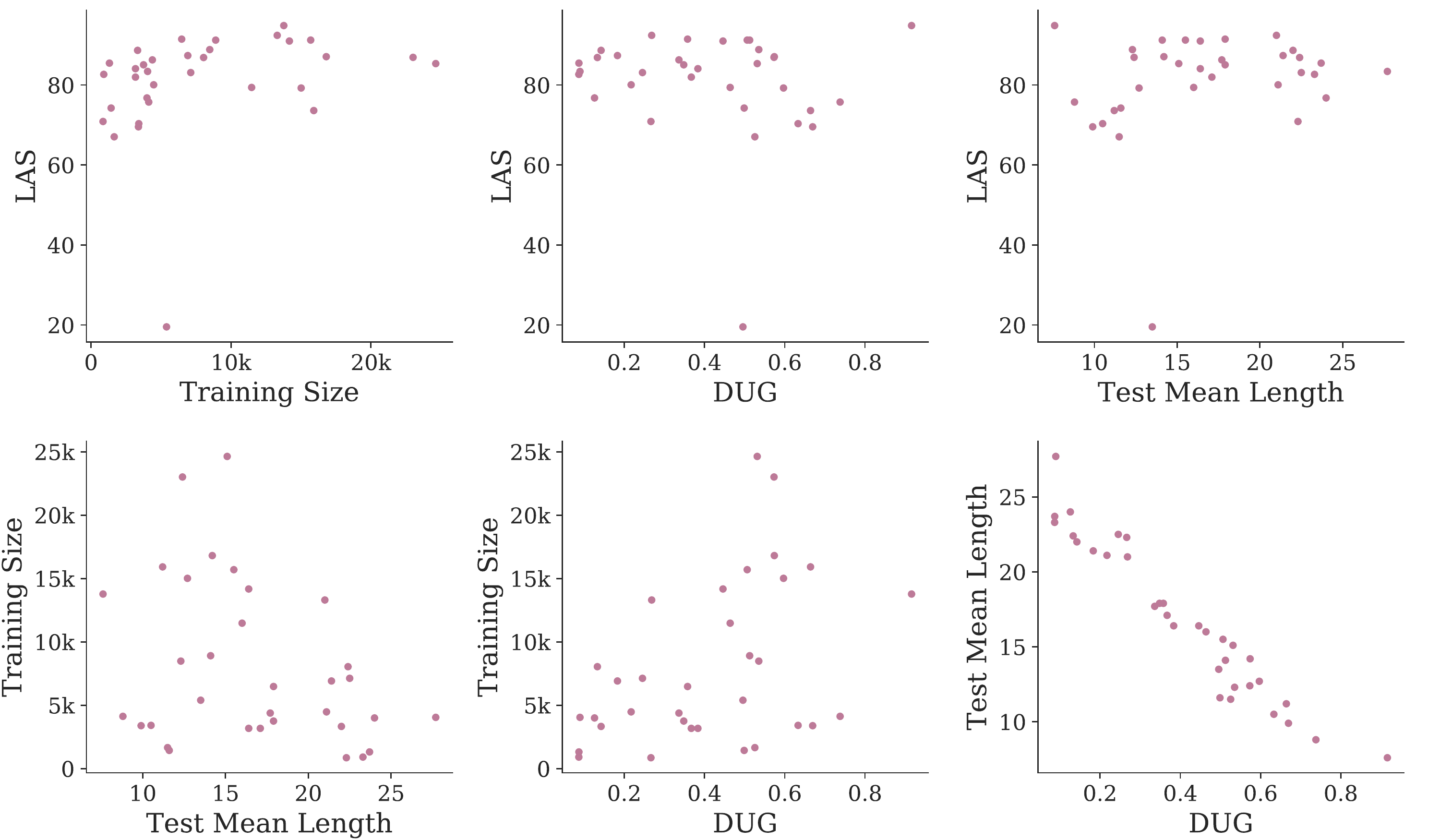}
    \caption{Data from original paper.}
    \label{fig:sogaard_scatters}
\end{subfigure} 

\begin{subfigure} 
    \centering
    \includegraphics[width=0.95\linewidth]{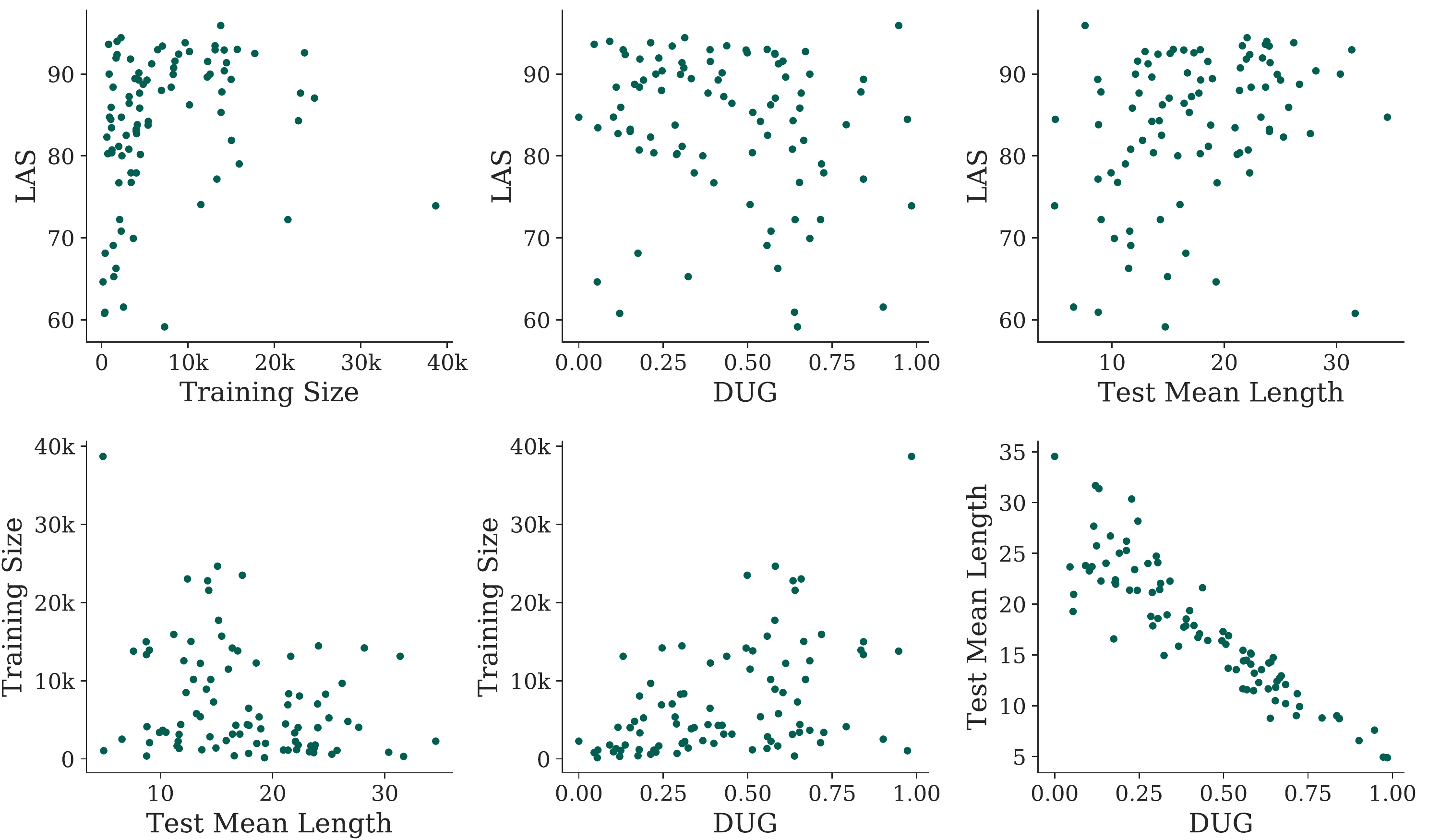}
    \caption{Data for UDPipe 2.0 and UD v2.7 using DUG.}
    \label{fig:udpipe2_scatters}
    \end{subfigure}     
\end{figure*}

\textcolor{black}{Figure \ref{fig:size_logged} shows the logarithmic relation between LAS and the training data size for UDPipe 2.0 and UD v2.7. Figure \ref{fig:sogaard_scatters} gives the visualisations for the data used in the original paper and Figure \ref{fig:udpipe2_scatters} gives the corresponding visualisation for UDPipe 2.0 and UD v2.7}.

\begin{figure*}[t]
    \centering
    \includegraphics[width=\linewidth]{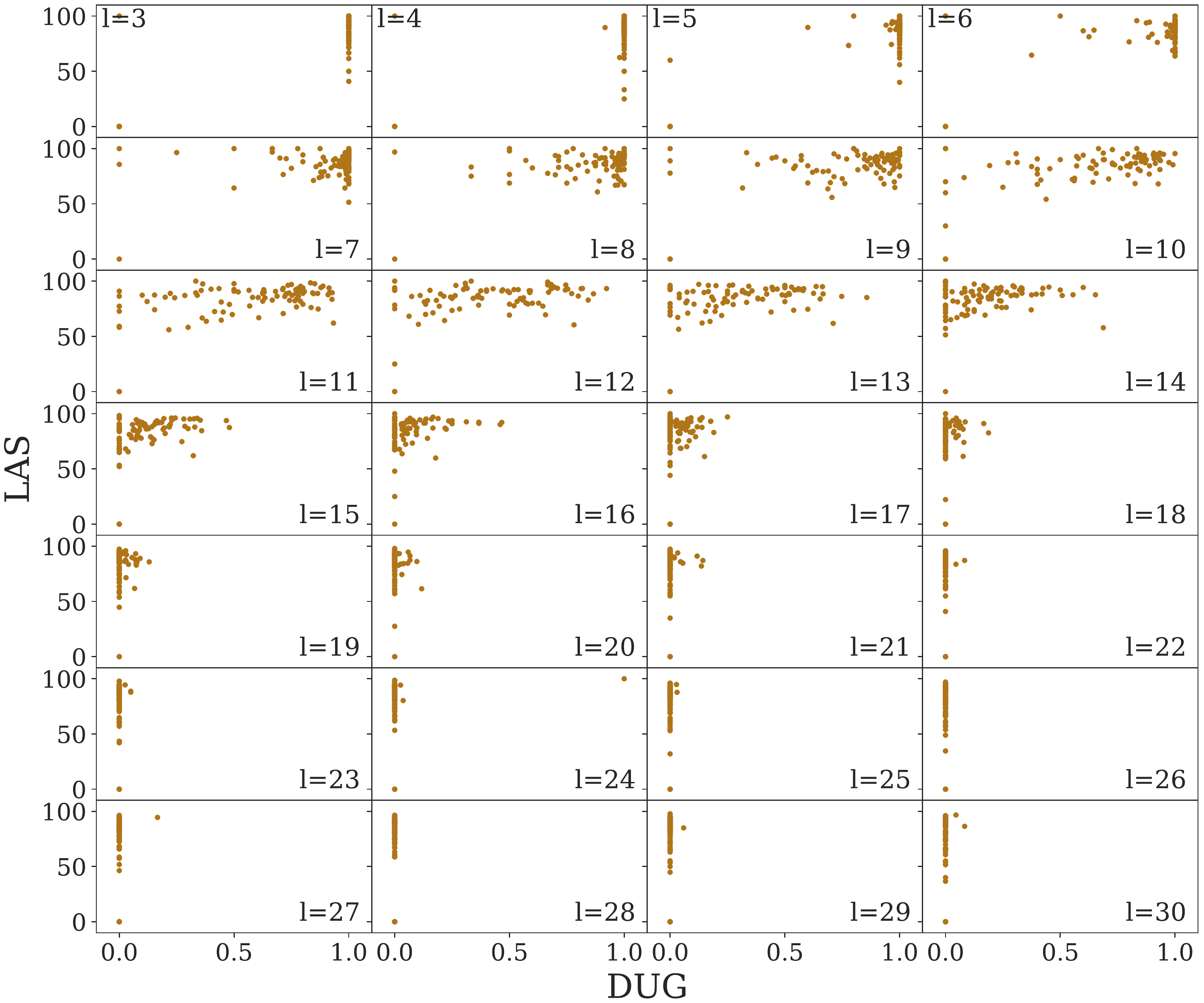}
    \caption{Length-binned analaysis. Data for UDPipe 2.0 and UD v2.7 using DUG.}
    \label{fig:dug-binned}
\end{figure*}

\begin{table}[tb!]
    \centering
    \small
    \begin{tabular}{lrr}
    \toprule
\multicolumn{1}{c}{} & \multicolumn{1}{c}{UDPipe 1.2} & \multicolumn{1}{c}{UDPipe 2.0}  \\
\midrule
\textbf{DUG} & 0.47 (p$<$0.001)  &  0.31 (p$=$0.003)   \\\midrule
\textbf{size}& -0.15 (p$=$0.153)  & -0.10 (p$=$0.335)  \\
\textbf{\MTL} & 0.64 (p$<$0.001) & 0.48 (p$<$0.001) \\
\textbf{both} & 0.17 (p$=$0.110) & 0.04 (p$=$0.683)  \\
\bottomrule
    \end{tabular}
    \caption{Partial Spearman's $\rho$ for focused DUG (i.e.\ using only the measurement for sentences of length 9 to 14) with covariants.}
    \label{tab:partial_correlations-focused}
\end{table}
\textcolor{black}{Figure \ref{fig:dug-binned} expands the example plots shown in Figure \ref{fig:example_sentence_length_bins} which only showed extreme cases. This shows LAS versus DUG for every sentence length bin from length 3 to 30. This clearly shows the issue with DUG as discussed in the main body.}

\textcolor{black}{All the data used for the analyses presented in this paper can be found in the supplementary material associated with the paper.}

\begin{table}[bh!]
    \centering
    \small
    \begin{tabular}{lrc}
    \toprule
    & \multicolumn{1}{c}{Spearman's $\rho$} & \multicolumn{1}{c}{p-value} \\
    \textbf{DUG}$\:\:$\textbf{LAS} & -0.184 & 0.083 \\
    \textbf{DUG}$\:\:$\textbf{LAS-bcg$_{\textrm{size}}$} &  -0.400 & 0.000 \\
    \textbf{DUG}$\:\:$\textbf{LAS-bcg$_{\textrm{size,Ltest}}$} & 0.010 & 0.926 \\\midrule
    $\langle L_{\textrm{Test}}\rangle\:\:$\textbf{LAS-bcg$_{\textrm{size}}$} &  0.465 & 0.000 \\
    \bottomrule
    \end{tabular}
    \caption{Correlation of DUG with LAS and then with LAS with the background associated with size and length (L) removed. Isolated row shows correlation of LAS without size background and mean sentence length in test data.}
    \label{tab:noise_correlations}
\end{table}

\end{document}